\documentclass[conference]{IEEEtran}

\usepackage{float}
\usepackage{footnote}
\usepackage{ucs}
\usepackage{amsmath}
\usepackage[utf8x]{inputenc}
\makesavenoteenv{tabular}
\usepackage{mdwmath}
\usepackage{mdwtab}

\ifCLASSINFOpdf
   \usepackage[pdftex]{graphicx}
\else
\fi

\begin{document}
%
\title{Automatic Language Identification for Romance Languages using Stop Words and Diacritics} 
\IEEEspecialpapernotice{Tool/Experimental Paper}

\author{
	\IEEEauthorblockN{Ciprian-Octavian Truic\u{a}}
	\IEEEauthorblockA{
		\small{Computer Science and Engineering Department}\\
		\small{Faculty of Automatic Control and Computers}\\
		\small{University Politehnica of Bucharest}\\
		\small{Bucharest, Romania}\\
		Email: ciprian.truica@cs.pub.ro}
	\and
	\IEEEauthorblockN{Julien Velcin}
	\IEEEauthorblockA{
		\small{University of Lyon}\\
		\small{ERIC Laboratory, Lyon 2}\\
		\small{Lyon, France}\\
		Email: julien.velcin@univ-lyon2.fr}
	\and
	\IEEEauthorblockN{Alexandru Boicea}
	\IEEEauthorblockA{
		\small{Computer Science and Engineering Department}\\
		\small{Faculty of Automatic Control and Computers}\\
		\small{University Politehnica of Bucharest}\\
		\small{Bucharest, Romania}\\
		Email: alexandru.boicea@cs.pub.ro}
}

\maketitle

\begin{abstract}
Automatic language identification is a natural language processing problem that tries to determine the natural language of a given content. In this paper we present a statistical method for automatic language identification of written text using dictionaries containing stop words and diacritics. We propose different approaches that combine the two dictionaries to accurately determine the language of textual corpora. This method was chosen because stop words and diacritics are very specific to a language, although some languages have some similar words and special characters they are not all common. The languages taken into account were romance languages because they are very similar and usually it is hard to distinguish between them from a computational point of view. We have tested our method using a Twitter corpus and a news article corpus. Both corpora consists of UTF-8 encoded text, so the diacritics could be taken into account, in the case that the text has no diacritics only the stop words are used to determine the language of the text. The experimental results show that the proposed method has an accuracy of over 90\% for small texts and over 99.8\% for large texts.
\end{abstract}

\vspace{-0.75em}
\IEEEpeerreviewmaketitle

\section{Introduction}

Automatic language identification (LID) is the problem of determining the natural language of a given content. As the textual data volume generated each day in different languages increases, LID is one of the basic steps needed for preprocessing text documents in order to do further textual analysis. \cite{IEEEhowto:CPeters2012}.

LID is one of the first basic text processing techniques used in Cross-language Information Retrieval (CLIR) - also known as Multilingual Information Retrieval (MLIR), text mining and knowledge discovery in text (KDT) \cite{IEEEhowto:RFeldman1995}. In these fields, the text processing step such as indexing, tokenization, part of speech tagging (POS), stemming and lemmatization are highly dependent on the language.

In this paper we propose a statistical method using two dictionaries for automatic language identification. The dictionaries contain stop words and diacritics (glyph added to a letter). The stop words dictionary is constructed using the most common words written with diacritics for the studied languages. Further more, we enhance this dictionary with terms written without diacritics. The diacritics dictionary contains the diacritics for each language. The method either use the dictionaries as they are, or weigh the terms based on the existence of the term in each language. To the best of our knowledge, these method has not been yet attempted in the literature.

The studied languages are romance languages, which are very similar to each other from a lexical point of view \cite{IEEEhowto:AMCiobanu2014} (Table~\ref{tab:table1}), this being the main reason why these languages were chosen for testing our statistical method.

\vspace{-.75em}
\begin{table}[H]
\centering
\caption{Lexical language similarities between the studied languages (N/A no data available)\protect\footnotemark}
\vspace{-0.75em}
\begin{tabular}{|l|r|r|r|r|r|}
\hline
\textbf{Language} & \multicolumn{1}{c|}{\textbf{French}} & \multicolumn{1}{c|}{\textbf{Italian}} & \multicolumn{1}{c|}{\textbf{Portuguese}} & \multicolumn{1}{c|}{\textbf{Romanian}} & \multicolumn{1}{c|}{\textbf{Spanish}} \\ \hline
\textbf{French} & 1 & 0.89 & 0.75 & 0.75 & 0.75 \\ \hline
\textbf{Italian} & 0.89 & 1 & N/A & 0.77 & 0.82 \\ \hline
\textbf{Portuguese} & 0.75 & N/A & 1 & 0.72 & 0.89 \\ \hline
\textbf{Romanian} & 0.75 & 0.77 & 0.72 & 1 & 0.71 \\ \hline
\textbf{Spanish} & 0.75 & 0.82 & 0.89 & 0.71 & 1 \\ \hline
\end{tabular}
\label{tab:table1}
\end{table}
\footnotetext{Ethnologue: Languages of the World http://www.ethnologue.com/}
\vspace{-1.5em}

This paper is structured as follows. Section II discusses related work done on automatic language detection. Section III presents our experimental setup and the proposed method for LID. Section IV presents experimental data and findings and discusses the results. The final section concludes with a discussion and a summary.

\vspace{-0.75em}
\section{Related work}

The major approaches for language identification include: detection based on stop words usage, detection based on character \emph{n-grams} frequency, detection based on machine learning (ML) and hybrid methods.

Stop words, from the point of view of LID, are defined as the most frequently terms as determined by a representative language sample. This list of words has been shown to be effective for language identification because these terms are very specific to a language \cite{IEEEhowto:CPeters2012}.

The method based on character \emph{n-grams} was first proposed by Ted Dunning \cite{IEEEhowto:TDunning1994}. To form \emph{n-grams}, the text drawn from a representative sample for each language is partitioned into overlapping sequences of \emph{n} characters. The counts of frequency from the resulting \emph{n-grams} form a characteristic \emph{fingerprint} of languages, which can be compared to a corresponding frequency analysis on the text for which the language is to be determined. In the literature there have been proposed applications and frameworks that identify the language based on this approach. They perform well on text collections extracted from Web Pages, but do not perform well for short texts \cite{IEEEhowto:SRajesh2013}. 

Xafopoulos et al. used the discrete  hidden Markov models (HMMs) with three models for comparison on a data set containing a collection of Web Pages collected over hotel Web Sites in 5 languages. The texts in the tested corpus were not noisy, because this genre of Web Sites presents the information as clean as possible \cite{IEEEhowto:AXafopoulos2004}. Their accuracy is between 95\% and 99\% for the sample corpus.

Timothy Baldwin et al. performed a study based on different classification methods using the nearest neighbor model with three distances, Naive Bayes and support vector machine (SVM) model \cite{IEEEhowto:TBaldwin2010}. Their results show that the nearest neighbor model using \emph{n-grams} is the most suitable for the tested datasets, with accuracy ranging form 87\% to 90\%. Other machine learning approaches use graph representation based on \emph{n-gram} features and graph similarities with an accuracy between 95\% and 98\% \cite{IEEEhowto:ETromp2011} or centroid-based classification with an accuracy between 97\% and 99.7\% \cite{IEEEhowto:HTakci2012}.

On short messages, like tweets, Bergsma et al. used Prediction by Pattern Matching and Logistic Regression with different features to classify languages under the same family of languages (Arabic, Slavic and Indo-Iranian) \cite{IEEEhowto:SBergsma2012}. Their accuracy is between 97\% and 98\% for the selected corpus.

Hybrid approaches use one or more of these methods. Abainia et al. propose different approaches that use term based and characters based methods \cite{IEEEhowto:KAbainia2014}. Their experiments show that the proposed hybrid methods are quite interesting and present good language identification performances in noisy texts, with accuracy ranging between 72\% and 97\%.

\vspace{-0.75em}
\section{Experimental setup and Algorithm}

The corpora used for experiments are freely available, and they can be downloaded from the Web\footnote{News articles and tweets corpora http://www.corpora.heliohost.org/}. For the experimental tests, a Twitter corpus and a news articles corpus are used. Before applying the method for LID, the texts' language was determined using the source of the data. This step is useful to determining the accuracy of our approach. Also, the texts are preprocessed to improve the language detection accuracy by removing punctuation and non-alphabetical characters.

Before applying the algorithm, the stop words and diacritics dictionaries were created. The stop words dictionary was downloaded from the Web\footnote{Stop words list http://www.ranks.nl/stopwords} and enhanced with stop words with no diacritics manually. This is useful because there are texts in the Twitter corpus written without diacritics and they can be misclassified. A total number of approximately 500 stop words is used for each language. 

\vspace{-0.75em}
\begin{table}[H]
\centering
\caption{Diacritics by language}
\vspace{-0.75em}
\begin{tabular}{|c|c|}
\hline
\textbf{Language} & \textbf{Diacritics} \\ \hline
French & àâæçèéêëîïôœùûü \\ \hline
Italian & àáèéìíòóùú \\ \hline
Portuguese & áâãàçéêíóôõú \\ \hline
Romanian & ăâîșşțţ \\ \hline
Spanish & áéíóúñü \\ \hline
\end{tabular}
\label{tab:table2}
\end{table}
\vspace{-1.5em}

Table~\ref{tab:table2} presents the diacritics used for LID. Although \emph{\c{t}} (t-cedilla) and \emph{\c{s}} (s-cedilla) are not Romanian diacritics, due to some wrong implementations of the correct diacritics \emph{\unichar{537}} (s-comma) and \emph{\unichar{539}} (t-comma), these were taken into account when constructing the diacritics dictionary.

The proposed method uses the term frequency of weighted stop words and diacritics. The term can be either a diacritic or a stop word. Using Equation (\ref{eq1}) each text receives a score based on the two dictionaries, the language that gets the highest score is selected as the language of the text. In case that there are no diacritics in the text the score will only be computed using the stop words dictionary, $p=1$. The final score is calculated as the weighted sum of the frequency of stop words plus the frequency of diacritics in the text for each language. 
The frequency can be computed using the two different measures presented in the previous chapter. 
The motivation for using a weight is based on the fact that the terms specific to a language should have a higher impact on the score than the ones that are common between languages, which is similar to the inverse document frequency (IDF) factor.

\vspace{-1.50em}
\begin{eqnarray}
\label{eq1}
& score(text, lang) = \nonumber \\
& p \cdot\sum_{w}f(w, text)\cdot weight(w, lang) + \\
& (1-p)\cdot\sum_{d}f(d, text)\cdot weight(d, lang) \nonumber
&  \nonumber \\
\label{eq2}
& f(term, text) = 
\begin{cases} 
	f_{term, text} \\
	\log(1 + f_{term, text})
\end{cases}\\
\label{eq3}
& weight(term, lang) = 
\begin{cases} 
	1 \\
	\frac{N}{n} \\
	\log(1+\frac{N}{n})
\end{cases}
\end{eqnarray}
\vspace{-1.25em}

In Equation (\ref{eq1}), the following notations are used: \emph{lang} represents the tested language, \emph{w} represents a stop word in the stop words dictionary for the given language, \emph{d} represents a diacritic in the dictionary of diacritics for the given language. Parameter \emph{$p \in [0, 1]$} is used to increase the impact of terms and to correlate stop words with diacritics.
The function \emph{$f(term, text)$} is used to compute the term frequency (Equation \ref{eq2}). It can be calculate as the raw term frequency for a term, the number of occurrences of a term in the text, \emph{$f(term, text)=f_{term, text}$}, or it can be normalized using the logarithmic normalization \emph{$f(term, text)=\log(1 + f_{term, text})$}. 
As the term frequency function, the term weighting function, \emph{$weight(term, lang)$}, can be computed using different approaches (Equation \ref{eq3}). If the terms are not weighted, then \emph{$weight(term, lang)=1$}. Another way for computing the weight is to take into account the frequency of a term in the studied languages. In this case, the fallowing formula for the weight function can be used \emph{$weight(term, lang) =\frac{N}{n}$}. In this formula \emph{N} is the number of languages tested and $n={|\{ term : term \in lang \}|}$ represents the number of languages that contains the term. The term $n$ can never be equal to $0$ because it appears at least in one language. As the term frequency function, the weight function can be logarithmic normalized \emph{$weight(term, lang)=\log(1+\frac{N}{n})$}. Although the last formula is similar to IDF from information retrieval (IR), it is not the same as it does not calculate the occurrence of a term in the corpus, but computes the occurrence of a dictionary term in the studied languages.

\vspace{-0.75em}
\section{Experimental Results}

The method is applied on two corpora. The first is a Twitter corpus containing 500\,000 tweets with 100\,000 tweets for each studied language. The second one is a news articles corpus that contains 250\,000 entries with 50\,000 articles for each studied language. All the text are encoded using UTF-8.  The accuracy of each approach is determined by using the initial classification of the texts. Although a small sample of the initial copora was manually validated, it should be noted that the initial classification was done automatically, based on the initial corpora sources, therefore the texts may contain noise or could be wrongly classified.

First we wanted to test the accuracy of our method whether we use only diacritics or only stop words.
Test 1 uses only diacritics, setting \emph{$p=0$} and no weight, \emph{$weight(term, lang)=1$}.
The second test (Test 2) uses weighted diacritics, setting \emph{$p=0$} and \emph{$weight(term, lang)=\frac{N}{n}$}.
For our third test (Test 3), we tested only stop words with no weight, setting \emph{$p=1$}. 
Test 4 deals with weighted stop words using \emph{$p=1$} and \emph{$weight(term, lang)=\frac{N}{n}$}.
The next two sets of tests use both diacritics and stop words with \emph{$weight(term, lang)=1$}.
For Test 5 we set \emph{$p=\frac{1}{2}$}.
Because diacritics are more language specific and they will better determine the language than stop words, we also wanted to test whether the accuracy improves if we set parameter \emph{p} to favor the second term of Equation (\ref{eq1}).
We choose to double the impact of diacritics for Test 6 by setting \emph{$p=\frac{1}{3}$}.
The next two sets of tests use weighted terms with \emph{$weight(term, lang)=\frac{N}{n}$}.
For Test 7 we set \emph{$p=\frac{1}{2}$}.
In Test 8 we enhanced the impact that diacritics have on determining the language to see if the accuracy improves further by setting \emph{$p=\frac{1}{3}$}.
For the last test (Test 9) we used the logarithmic normalization for the term frequency, \emph{$f(term, text)=\log(1 + f_{term, text})$}, and the weight, \emph{$weight(term, lang)=\log(1+\frac{N}{n})$}, together with $p=\frac{1}{3}$.

Table~\ref{tab:table3} presents the experimental result of the method applied on the Twitter corpus and Table~\ref{tab:table4} presents the results of the method applied to the news article corpus.

\vspace{-0.75em}
\begin{table}[H]
\centering
\caption{Twitter corpus accuracy comparison}
\label{tab:table3}
\vspace{-0.75em}
\begin{tabular}{|c|c|c|c|c|c|}
\hline
{\bf Test No.} & {\bf French}  & {\bf Italian} & {\bf Portuguese} & {\bf Romanian} & {\bf Spanish} \\ \hline
Test 1         & 6.31\%        & 8.67\%        & 34.53\%          & 22.39\%        & 12.49\%       \\ \hline
Test 2         & 6.31\%        & 8.69\%        & 34.54\%          & 22.40\%        & 12.57\%       \\ \hline
Test 3         & 92.80\%       & 90.78\%       & 88.79\%          & 83.00\%        & 89.49\%       \\ \hline
Test 4         & 93.14\%       & 90.98\%       & 88.95\%          & 88.59\%        & 90.46\%       \\ \hline
Test 5         & 93.62\%       & 91.33\%       & 90.62\%          & 85.40\%        & 90.68\%       \\ \hline
Test 6         & 93.64\%       & 91.29\%       & 90.70\%          & 85.60\%        & 90.73\%       \\ \hline
Test 7         & 93.88\%       & 91.57\%       & 90.70\%          & 90.15\%        & 91.59\%       \\ \hline
Test 8         & 93.90\%       & 91.56\%       & 90.78\%          & {\bf 90.18\%}  & 91.63\%       \\ \hline
Test 9         & {\bf 94.09\%} & {\bf 91.68\%} & {\bf 91.02\%}    & 90.07\%        & {\bf 92.18\%} \\ \hline
\end{tabular}
\end{table}

\vspace{-2.0em}
\begin{figure}[H]
\centering
\includegraphics[width=3.2in]{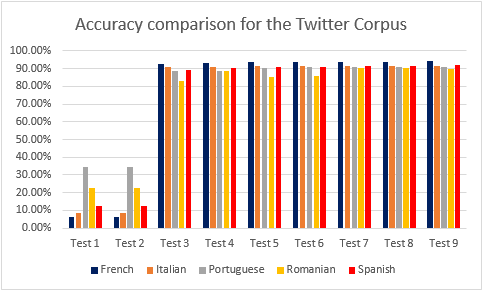}
\vspace{-0.75em}
\caption{Accuracy comparison for the Twitter Corpus}
\label{fig1}
\end{figure}
\vspace{-1.5em}

When using weighted stop words and diacritics with our method the accuracy is over 90\% for the twitter corpus for all the tested languages. From our test results, we can conclude that the highest accuracy is achieved when \emph{$p=\frac{1}{3}$} for weighted term, showing the real impact of diacritics for LID. In Figure \ref{fig1} we compare the accuracy of each test for each language.

\begin{figure}[H] 
\centering
\includegraphics[width=3.2in]{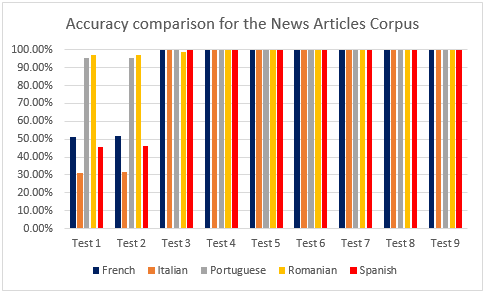}
\vspace{-0.75em}
\caption{Accuracy comparison for the News Articles Corpus}
\label{fig2}
\end{figure}

From the experimental results, one can conclude that by setting \emph{$p=\frac{1}{2}$} and \emph{$p=\frac{1}{3}$}, work very well on the news article corpus with an almost 100\% accuracy for all the tested languages, the lowest accuracy being over 99.8\%. Text size has a significant impact on the method because more information is available for classification. Adding weights to stop words and diacritics proves to be a very efficient approach for the Romanian and Italian languages. Taking into account only stop words (Test 3 and Test 4) yields an accuracy of over 99\% for automatically determining the languages of news articles. The method's accuracy is high (Figure \ref{fig2}) because the information in this corpus does not present a lot of noise.

\vspace{-0.75em}
\begin{table}[H]
\centering
\caption{News articles corpus accuracy comparison}
\label{tab:table4}
\vspace{-0.5em}
\begin{tabular}{|c|c|c|c|c|c|}
\hline
{\bf Test No.} & {\bf French}  & {\bf Italian} & {\bf Portuguese} & {\bf Romanian} & {\bf Spanish} \\ \hline
Test 1         & 51.49\%       & 31.34\%       & 95.56\%          & 96.90\%       & 45.54\%       \\ \hline
Test 2         & 51.51\%       & 31.42\%       & 95.57\%          & 96.89\%       & 46.02\%       \\ \hline
Test 3         & 99.89\%       & 99.77\%       & 99.86\%          & 99.03\%       & 99.95\%       \\ \hline
Test 4         & 99.89\%       & 99.82\%       & 99.86\%          & 99.78\%       & 99.95\%       \\ \hline
Test 5         & 99.92\%       & 99.80\%       & 99.97\%          & 99.91\%       & 99.96\%       \\ \hline
Test 6         & {\bf 99.93\%} & 99.80\%       & {\bf 99.98\%}    & 99.93\%       & 99.96\%       \\ \hline
Test 7         & 99.92\%       & {\bf 99.84\%} & 99.97\%          & {\bf 99.97\%} & 99.95\%       \\ \hline
Test 8         & {\bf 99.93\%} & {\bf 99.84\%} & 99.97\%          & {\bf 99.97\%} & 99.96\%       \\ \hline
Test 9         & {\bf 99.93\%} & {\bf 99.84\%} & {\bf 99.98\%}    & {\bf 99.97\%} & {\bf 99.97\%} \\ \hline
\end{tabular}
\end{table}

\vspace{-2em}
\vspace{-1.5em}

Table \ref{tab:table5} show the wrongly classified tweets by Test 9, where we set \emph{$p=\frac{1}{3}$}, $f(term, text)=\log(1+f_{term, text})$ and \emph{$weight(term, lang)=\log(1+\frac{N}{n})$}. This miss-classification can be attributed to a high percentage of tweets, between 5\% and 8\%, that could not be classified. This set of tests has the overall lowest percentage of tweets that could not be classified.

\vspace{-0.75em}
\begin{table}[H] 
\centering
\label{tab:table5}
\caption{Tweets wrongly classified by Test 9}
\vspace{-0.75em}
\begin{tabular}{|c|c|c|c|c|c|}
\hline
{\bf Language} & {\bf French} & {\bf Italian} & {\bf Portuguese} & {\bf Romanian} & {\bf Spanish} \\ \hline
French         & 94.09\%      & 0.32\%        & 0.26\%           & 0.69\%        & 0.30\%        \\ \hline
Italian        & 0.22\%       & 91.68\%       & 0.75\%           & 1.12\%        & 0.63\%        \\ \hline
Portuguese     & 0.20\%       & 0.19\%        & 91.02\%          & 0.25\%        & 1.03\%        \\ \hline
Romanian        & 0.03\%       & 0.06\%        & 0.09\%           & 90.07\%       & 0.07\%        \\ \hline
Spanish        & 0.13\%       & 0.14\%        & 0.37\%           & 0.17\%        & 92.18\%       \\ \hline
Not Classified & {\bf 5.33\%} & {\bf 7.61\%}  & {\bf 7.51\%}     & {\bf 7.70\%}  & {\bf 5.79\%}  \\ \hline
\end{tabular}
\end{table}
\vspace{-1.5em}

As expected, the accuracy is lower when taking into account only stop words or diacritics than when both dictionaries are used for LID. The accuracy discrepancy between the different sets of tests can be seen especially for the Twitter corpus where, for some languages, the difference between accuracies are considerably high (Figure \ref{fig1}). Instead, for the news articles corpus this approach achieves an accuracy of over 96\% for Romanian and Portuguese. This could be attributed to the fact that Romanian and Portuguese have more diacritics that are widely used and are not found in the other studied languages. 

Miss-classification appears because diacritics and stop words are common between languages. For example the Spanish tweet \emph{allí estaré} cannot be classified because our method will only compute the score based on the diacritics \emph{í} and \emph{é}, which appear in Italian, Spanish and Portuguese. The French tweet \emph{il y a plongé son visage} is wrongly classified as Spanish. This miss-classification arises because the term \emph{y} appears as a Spanish stop word but not as a French one. The Italian tweet \emph{buona sera wagliù} is miss-classified as French. In this case the term \emph{sera} is considered as a French stop word but not an Italian one. The Romanian tweet \emph{universitate facultate istorie} cannot be classified because it does not contain stop words or diacritics. The same miss-classification appears for the Italian tweet \emph{messaggio ricevuto}. The dictionaries must be carefully built so that miss-classifications do not appear. 

\vspace{-0.75em}
\section{Conclusion}

Stop words prove to be very effective for automatic language identification. Although with different semantical meanings, stop words can be very similar, even the same, for languages that are related. For example the word \emph{la} appears as a word in French, Italian, Spanish and Romanian, but does not appear in Portuguese. Other terms are very specific to a language and the scoring can be skipped altogether if they are found, e.g., the stop word \emph{și} (s-comma and i) is only found in Romanian, \emph{votre} is only found in French, etc.

Diacritics also prove very useful for LID. Some diacritics appear for certain language, e.g., \emph{ț} (t-comma) for Romanian, \emph{ì} (i with grave accent) for Italian, \emph{ã} (a-tilde) for Portuguese, etc. If the analyzed text is written correctly, then, only by looking at the set of diacritics and the stop words, a LID method can accurately classify the given text and no scoring is necessary, especially in the cases where the diacritics are unique to the language and are widely used.

The experimental results show that for the news articles corpus a very high accuracy is achieved when using the method with both dictionaries to classify the corpus, almost all the texts are correctly classified with an accuracy of over 99.8\%. This result can be attributed to the fact that the texts from the news articles are written using diacritics. The size of the text has a considerable impact on the accuracy as it present more information to the method.

For the Twitter corpus the method using weighted terns with \emph{$p=\frac{1}{3}$} achieves the overall best accuracy. Although, the accuracy difference between Test 7 and 8 is very small, almost insignificant, under 0.1\%. The best accuracy is achieved when the logarithmic normalization is used for the term frequency and the weight (Text 9). To improve the accuracy for short texts and lower miss-classification, our method can be used together with the already existing major approaches for LID.

It should be taken into account that these languages come from the same linguistic family and they are very similar from a lexical point of view, and some of the diacritics and stop words are common. The accuracy of the Twitter corpus is lower than the accuracy of the news articles corpus because not all the texts are written using diacritics. Some noise could also be found in this corpus, some tweets could only contain links or hashtags, which were not taken into account when doing LID.

The presented method is very flexible and it can be implemented in any programming language and at any application layer. It does not need any training as it is needed for LID algorithms that use ML, being highly dependent on the natural language instead of a training corpus. There is a minimal number of computations done, the temporal complexity being $O(|w|+|d|)$, where $|w|$ is equal to the number of stop words and $|d|$ is the equal to the number of diacritics used. Because this method is based on dictionaries, the score function can upgrade or degrade to a different one, giving flexibility when one of the dictionaries is missing.

In conclusion, although stop words usually are a good measure for automatic language detection, for small and large texts the accuracy of LID can be improved by using diacritics. Based on the experimental result, we can observe that if we want to improve accuracy and remove miss-classification the stop words dictionary must be well-built, conveniently by experts in the field.

In future work, we seek to fine tune the parameter $p$ by automatically learning it to see if we can achieve better accuracy. We also want to test this statistical approach on other Romance languages (e.g. Occitan, Catalan, Venetian, Aromanian, Galician etc.) and other Indo-European families of languages such as Germanic and Slavic. From an architectural point of view, we also want to parallelize the algorithm as much as possible to reach real-time performance.

\end{document}